\documentclass[conference]{IEEEtran}

\usepackage{amsmath}
\usepackage{graphicx}
\usepackage[hidelinks]{hyperref}
\usepackage{algpseudocode}
\usepackage{algorithm}
\usepackage{xcolor}
\usepackage{ulem}
\usepackage{multirow}
\usepackage{cite}
\usepackage{tikz}
\usepackage{hyperref} 

\begin{document}

\title{On-Device Neural Architecture Search}

\author{\IEEEauthorblockN{Andrea~Mattia~Garavagno\IEEEauthorrefmark{1}\IEEEauthorrefmark{2}\IEEEauthorrefmark{3},
        Edoardo~Ragusa\IEEEauthorrefmark{1},
        Paolo~Gastaldo\IEEEauthorrefmark{1},
        Antonio~Frisoli\IEEEauthorrefmark{2}\IEEEauthorrefmark{3},
        and~Claudio~Loconsole\IEEEauthorrefmark{4}\IEEEauthorrefmark{3}
        \IEEEauthorblockA{\IEEEauthorrefmark{1} Department of Electrical, Electronic, Telecommunication Engineering and Naval Architecture, DITEN \\ University of Genoa, Genoa 16145, Italy}
        \IEEEauthorblockA{\IEEEauthorrefmark{2} Department of Excellence in Robotics \& AI, Scuola Superiore Sant’Anna, Piazza Martiri della Libertà 33, Pisa 56127, Italy \\}
        \IEEEauthorblockA{\IEEEauthorrefmark{3} Institute of Mechanical Intelligence, Scuola Superiore Sant’Anna, Ghezzano, 56010 Pisa, Italy}
        \IEEEauthorblockA{\IEEEauthorrefmark{4} Faculty of Technological and Innovation Sciences, Universitas Mercatorum, 00186 Rome, Italy \\ e-mail: AndreaMattia.Garavagno@\{edu.unige.it, santannapisa.it\}}}
}

\IEEEoverridecommandlockouts


\maketitle

\begin{tikzpicture}[remember picture,overlay]
\node[anchor=south,yshift=15pt] at (current page.south) {
    \parbox{\textwidth}{
        \centering \scriptsize
        This work has been accepted for publication in the proceedings of the 2025 IEEE International Conference on Consumer Electronics (ICCE). \copyright~2025 IEEE. Personal use of this material is permitted.  Permission from IEEE must be obtained for all other uses, in any current or future media, including reprinting/republishing this material for advertising or promotional purposes, creating new collective works, for resale or redistribution to servers or lists, or reuse of any copyrighted component of this work in other works. Final publication is available at \url{https://doi.org/10.1109/ICCE63647.2025.10929801}.
    }
};
\end{tikzpicture}

\IEEEpubidadjcol

\begin{abstract}
This paper proposes a new approach to near-sensor computing, in which a lightweight Neural Architecture Search (NAS) is performed directly on the deployment device to find the best tiny neural architecture for analyzing the real-time data acquired through sensors.
This new adaptation capability can be particularly useful in the case of human-machine interfaces for which the neural network analyzing the biometrical data can be re-designed each time the user changes, after a guided data collection procedure, fighting the typical data variations between individuals on a new level. 
To implement the proposed approach a new NAS has been designed and then validated on the Italian Sign Language dataset (ISL), a collection of surface electromyography (sEMG) signals of the signs of the Italian alphabet, using several embedded systems. Moreover, further validation on the Case Western Reserve University dataset (CWRU), a benchmark for intelligent fault diagnosis, is presented to suggest another possible application of the proposed approach. 
When run on a Raspberry Pi 4, the proposed NAS performs beyond the state of the art proposing a tiny neural architecture having 0.63 times less RAM occupancy and 5.96 percentage points of more accuracy in the case of the ISL dataset; and 0.44 times less RAM occupancy and 0.2 percentage points of more accuracy in the case of the CWRU dataset.
\end{abstract}


\IEEEpeerreviewmaketitle

\section{Introduction}\label{sec:introduction}
The present work proposes a new approach to near-sensor computing \cite{zhou2020near}, in which a lightweight Neural Architecture Search (NAS) is performed directly on the deployment device to find the best tiny neural architecture for analyzing the real-time data acquired through sensors. In this scenario, the sensor node can collect labelled data from human feedback, find a tiny neural architecture tailored to the collected data thanks to a lightweight search procedure, and then use it to analyze the data from the attached sensors in real-time. This novel near-sensor computing approach has the potential to enhance continual learning \cite{hadsell2020embracing}, allowing not only weights but also the architecture itself to co-adapt with the data drift. The new adaptive capability has the potential to enhance a next-generation of Human-Machine Interfaces (HMIs) that analyze real-time data from biometric sensors. Each time the user changes, the tiny neural architecture analyzing biometric data is redesigned after a guided data collection procedure, addressing the typical data variations between individuals \cite{ortega2004authentication} on a new level. 

The same approach could support applications typical of near-sensor computing such as e-health, personalized advertisement, structural health monitoring, and intelligent fault diagnosis. 
In the latter use case, an application-agnostic intelligent fault diagnosis system (IFDS) could be designed in-house, using the proposed approach, and then deployed on the field. Personnel could manually tamper the machine to be monitored to gather fault data with the available sensors; then collected data feed the application-agnostic IFDS, which automatically designs a custom neural architecture for classifying faults in real-time. In this way, a single IFDS can satisfy the requirements of multiple applications, as the neural architecture can adapt to the specific set of sensors used to acquire fault data. 

\section{Related Work}
The quest towards implementing deep learning on edge devices started with the inference phase \cite{jacob2018quantization, lin2020mcunet, lin2021mcunetv2}, proceeded with the training phase \cite{lin2022device, lee2021overview, yang2023efficient}, and is leaning towards the design phase \cite{garavagno2024running, garavagno2024affordable} given the recent results provided by hardware-aware NAS (HW-NAS) \cite{ragusa2024combining, risso2022lightweight, rala2021neural, zhang2020fast}. 

Garavagno et al. \cite{garavagno2024running} reformulated the optimization problem of HW-NAS by including in the set of constraints the resources available in the device performing the search procedure. Such constraints add to the constraints imposed by the deployment device, which usually characterize the classical versions of HW-NAS; this in turn further reduces the search space. The goal is to avoid the evaluation of architectures not fitting the resources available in the device performing the search procedure; as such, one can enable practical implementation of HW-NAS on consumer devices. In \cite{garavagno2024affordable}, Garavagno et al. further developed the idea of running the search procedure on embedded devices, proposing an adaptive search strategy, which tunes the length of the evaluation process according to the available computational resources.

The objective of \cite{garavagno2024running} and \cite{garavagno2024affordable} was to automatically design custom neural architectures inside a gateway, one for each type of sensor node connected to it. Such approach preserves privacy, as each device cab exploit the locally acquired data, without involving any third party. Conversely, this paper aims to propose a different approach, where the NAS is expected to run on the deployment device. Another main difference is that the NAS implemented in \cite{garavagno2024running} and \cite{garavagno2024affordable} produce models meant to analyze images, while the NAS proposed in the present document produces models meant to analyze time series. 

\section{Proposed Neural Architecture Search}
\subsection{Search Space}
The proposed NAS organizes the input time series from different sensors as a bi-dimensional matrix where each row contains the samples belonging to one sensor. Hence obtaining a matrix having $n_{sensors}$ rows and $n_{samples}$ columns. In such a way the resulting matrix can be analyzed by a bi-dimensional convolutional neural network built according to the following rules: the input is a base cell, composed of a bi-dimensional convolutional layer having $k$ kernels, followed by a batch normalization layer and a ReLu activation. Then $c$ cells, composed of a bi-dimensional max pooling layer followed by a bi-dimensional convolutional layer, a batch normalization layer, and a ReLu activation, are stacked upon the base cell. The number of kernels used in each cell is computed using the formula presented in (\ref{eq:kernels}). The extracted features are then reduced by a global average pooling layer and finally classified by a fully connected layer, having Softmax as the activation function.

\begin{equation} \label{eq:kernels}
\begin{aligned}
n_{c} = \begin{cases}
   \multicolumn{1}{@{}c@{\quad}}{k} & if \quad c = 0\\
   \left \lceil{ (2 - \sum_{i=1} ^{c - 1} 2^{-i}) \cdot n_{c - 1} }\right \rceil & if \quad c \geq 1
\end{cases}
\end{aligned}
\end{equation} 

To preserve the input matrix height, kernels of $n_{sensors}$ height are used, and stride 1 is always applied to the height dimension. Instead, on the rows dimension, a stride equal to two and a kernel as wide as two samples are always applied in the case of pooling layers, this is done to halve the time series length each time, whereas, in the case of convolution, a stride equal to one is applied alongside a kernel width of three. All the convolutional layers employ zero padding to maintain the input's resolution intact. 

\subsection{Optimization Problem}\label{sec:optimization_problem}
Classical HW-NAS aims to find the best neural architecture for the target hardware in the search space. This leads to a constrained optimization problem where the objective function describes the metric adopted to evaluate candidate solutions and the constraints set the boundaries of the search according to the deployment hardware chosen. Garavagno et al. \cite{garavagno2024running} proposed to include in the constraints the  resources available on the hardware performing the search, which usually were considered unbounded. As a result, HW-NAS can run on a resource-constrained device. 

In the case of the approach proposed in the present paper, the device performing the search procedure coincides with the device used for deployment. Therefore, the constraints refer to a single target hardware.  Indeed, as the goal is to deal with real-time applications (see  Sec. \ref{sec:introduction}), a constraint on the inference time is also added. The network's input size is considered fixed during the search since it depends on the user's needs. Hence, the search variable $x = (k, c)$ is defined without including the network's input size. This leads to problem formulation $P$ 

\begin{equation} \label{eq:problem}
\begin{aligned}
P: \begin{cases}
   \hfil \max f(x)\\
   \phi_{M}(x) \leq \xi_{M} , \phi_{T}(x) \leq \xi_{T} \\
   \hfil \xi_{M}, \xi_{T} > 0
\end{cases}
\end{aligned}
\end{equation}

where function $f$ returns the maximum validation accuracy obtained during the evaluation phase, function $\phi_{M}$ returns the memory occupancy during training and function $\phi_{T}$ returns the time consumed by one inference. 
Parameters $\xi_{M}$ and $\xi_{T}$ respectively represent the upper bounds for the candidate's memory occupancy during training and time per inference.

\subsection{Search Strategy}
In the proposed approach the developed NAS runs on constrained devices; thus, there is a need for a lightweight search strategy. Accordingly, the derivative-free search strategy proposed in \cite{garavagno2024colabnas} has been reformulated to solve the optimization problem presented in section \ref{sec:optimization_problem}. In detail, the phase in which a candidate network is evaluated has been reorganized as a child process. As such, its parent can monitor if the child process has been killed by the operative system because of memory depletion; this in turn means that the candidate network violated the constraint on the available memory. Moreover, the evaluation phase now measures the inference time of the candidate network to determine if such solution satisfies the time constraint. 

\section{Experimental Validation}
To provide experimental validation of the proposed approach, the developed NAS has been run on three different embedded devices, namely: the Raspberry Pi 4, the Raspberry Pi 3, and the Raspberry Pi Zero 2 W, which are often used in wearable and intelligent systems \cite{mathe2024comprehensive}. Table \ref{tab:embedded_boards} provides the hardware specifications for each device.

\begin{table}[!h]
    \centering
        \begin{tabular}{c c c c} 
                        \bf{Embedded Device} & \bf{SoC}      & \bf{RAM}   \tabularnewline 
                        \bf{Raspberry Pi}    & \bf{Broadcom} & \bf{[GiB]} \tabularnewline \hline
                        4 Model B            & BCM2711       & 4          \tabularnewline 
                        3 Model B            & BCM2837       & 1          \tabularnewline 
                        Zero 2 W             & BCM2710A1     & 0.5        \tabularnewline
        \end{tabular} 
        \caption{Specifications of the embedded devices targeted by this work.}
        \label{tab:embedded_boards} 
\end{table}

\begin{table*}[!ht]
    \centering
    \begin{tabular}{c | c | c c | c c c c c || c c c c c}
       \multirow{3}{*}{dataset} & \multirow{3}{*}{device} & \multicolumn{2}{c}{search cost} & \multicolumn{5}{|c||}{resulting architecture}                              & \multicolumn{5}{c}{reference architecture}  \\ 
                                &                         & time      & energy              & \multirow{2}{*}{(k, c)} & RAM    & Flash     & test acc. & latency         & \multirow{2}{*}{work} & RAM    & Flash     & test acc. & latency \\ 
                                &                         & [hh]:[mm] & [Wh]                &                         & [kiB]  & [kiB]     & [\%]      & [ms]            &                       & [kiB]  & [kiB]     & [\%]      & [ms]    \\ \hline
       \multirow{3}{*}{ISL}     & RPi 4                   & 8:11      & 37.2                & (16, 4)                 & 81.5   & 224.5     & 99.9\%    & $15.3 \pm 0.18$ & \multirow{3}{*}{\cite{pau2023electromyography}}     & \multirow{3}{*}{128.5} & \multirow{3}{*}{96.7}  & \multirow{3}{*}{93.94} & $0.6 \pm 0.04$  \\
                                & RPi 3                   & 14:42     & 54.4                & (8, 4)                  & 42.5   & 62.5      & 92.4\%    & $10.2 \pm 0.4$  &                       &       &           &           & $1.2 \pm 0.08$         \\
                                & RPi Zero 2              & 12:03     & 23.9                & (4,0)                   & 17.5   & 2.7       & 44.3\%    & $1.2 \pm 0.34$  &                       &       &           &           & $1.3 \pm 0.3$    \\ \hline
       \multirow{3}{*}{CWRU}    & RPi 4                   & 3:42      & 14.6                & (16, 4)                 & 29.5   & 39.3      & 99.5\%    & $0.7 \pm 0.01$  & \multirow{3}{*}{\cite{chen2020improved}}     & \multirow{3}{*}{66.5} & \multirow{3}{*}{163.39} & \multirow{3}{*}{99.3} & $0.2 \pm 0.02$  \\
                                & RPi 3                   & 7:58      & 27.9                & (16, 3)                 & 28.5   & 24.5      & 99.2\%    & $1.1 \pm 0.04$  &                       &        &           &           & $0.5 \pm 0.05$        \\ 
                                & RPi Zero 2              & 5:03      & 10.2                & (4, 4)                  & 10     & 9.45      & 94.6\%    & $0.4 \pm 0.17$  &                       &        &           &           & $0.6 \pm 0.23$        \\
    \end{tabular}
    \caption{Resulting Architectures VS Reference Architectures}
    \label{tab:results}
\end{table*} 
Two datasets have been selected according to the use cases proposed in section \ref{sec:introduction}: the Italian Sign Language (ISL) dataset \cite{sernani2021italian} to demonstrate that the proposed approach can be effectively used to automatically design neural architectures for human-machine interfaces, and the Case Western Reserve University (CWRU) dataset \cite{smith2015rolling} to demonstrate that the proposed approach can be also applied to design neural architectures for intelligent fault diagnosis systems. 

The ISL dataset \cite{sernani2021italian} contains surface electromyography (sEMG) and inertial measurements regarding the 26 letters of the hand sign Italian alphabet. 8 electrodes of the Myo Gesture Control Armband have been used to acquire the sEMG data at 200 Hz for 2 seconds. 30 samples for each sign have been collected, for a total of 780. To compare the resulting architecture with the one proposed by Pau et al. \cite{pau2023electromyography}, only the sEMG data regarding A, C, H, J, K, M, N, P, S, W, X signs were used, for a total of 11 classes and 330 samples. The test split of 20\% which brought the best results in terms of accuracy score in \cite{pau2023electromyography} has been adopted. 

The CWRU dataset \cite{smith2015rolling} contains vibration data for normal and faulty bearings.
To compare the resulting architecture with the work proposed by Chen et al. \cite{chen2020improved}, the vibration data from the acceleration sensor at the driving end were selected at 12 kHz sampling frequency. Absence of faults and 9 types of faults, namely ball and inner raceway at 0.007, 0.014, 0.021 inches, and outer raceway oriented at 3, 6, and 12 o'clock at 0.007 inches were selected at 1797 rpm, i.e. at 0 HP. 100 of samples of length 1024 are collected for each class, for a total of 1000 samples.

In both cases, an upper bound of 0.33 seconds, equivalent to 30 inferences per second has been set to the latency, as the constraint of real-time. All models are trained using quantization-aware training and then quantized to 8-bit. Architectures are evaluated for $500$ epochs with a learning rate of $10^{-3}$ and a batch size of $256$.

Table \ref{tab:results} compares the resulting architectures with the state-of-the-art work for each dataset, and also reports the time and the energy needed by the NAS to obtain the result. RAM and Flash occupancies have been measured using X-CUBE-AI's stm32tflm executable, which reports resource usage when using Tensorflow Lite for Microcontrollers as runtime. Whereas, latency is measured by running the model 1000 times on the chosen hardware using Tensorflow Lite runtime. 

As can be seen, the resulting architectures' footprint decrease according to the resources available to the embedded device performing the search. The main cause is the decrease in the memory available for training, which further restricts the search space, not allowing the algorithm to explore solutions with larger footprints. The latter decrease also causes a loss in test accuracy. However, it does not cause an increase in the latency: even if the processor's computational power decreases, the footprint reduction overcomes the decrease, providing lower latencies. 
In any case, the proposed approach provides significantly smaller footprints than the reference work for both datasets, whatever execution platform is considered. Still, the test accuracy provided surpasses the one offered by the reference works in the case of the Raspberry Pi 4 as an execution platform while remaining comparable in the case of the Raspberry Pi 3. Only in the case of Raspberry Pi Zero 2, the accuracy score obtained is drastically lower, especially in the case of the ISL dataset, in which no candidate architecture provides satisfactory results within such a tight memory constraint for training. Even if the total power consumption could be provided by a common laptop battery, the search cost of the proposed approach is high, suggesting a run overnight.

\section{Conclusion}
The present document proposed a new approach to NAS where the search procedure is performed directly on the deployment device. This novel approach enables a new paradigm where an edge device can collect labelled data from human feedback, find a neural architecture tailored to the collected data, and use it to perform inference in real time. The latter paradigm has the potential to enhance continual learning, allowing not only weights but also architectural changes to co-adapt with the data drift.

The proposed approach has been validated on two different datasets: the Italian Sign Language dataset and the Case Western Reserve University dataset; using three different embedded devices: the Raspberry Pi 4 Mobel B, the Raspberry Pi 3 Mobel B, and the Raspberry Pi Zero 2 W; demonstrating a promising capacity of delivering state-of-the-art architectures not only in the domain of biometrical signals for human-machine interfaces, but also in the domain of intelligent fault diagnosis. 

The search cost measured during experiments suggests overnight runs. In the latter scenario, the human-machine interface can adapt to the new owner overnight, delivering a tailored neural architecture for the real-time analysis of the user's biometrical signals. However, considering that the amount of data collected from a single user would be less than the one of the datasets adopted as benchmark, the search cost could be lower. For the latter use case, further experiments are needed the asses the real search cost.

\section*{Acknowledgment}
\addcontentsline{toc}{section}{Acknowledgment}
\scriptsize
Project funded under the National Recovery and Resilience Plan (NRRP), Mission 4 Component 2 Investment 1.1 - Call for tender No. 1409 published on Sept 14, 2022 by the Italian Ministry of University and Research (MUR) funded by the European Union – NextGenerationEU - Project Title "LEARN - muLtimodal Edge computing-bAsed
weaRable exoskeletoNs for assistance in daily life" – CUP: D53D23016190001, D53D23016200001, J53D23014090001 - Grant Assignment Decree No. 1383 adopted on September 01, 2023 by the Italian Ministry of University and Research (MUR).

\vspace{5pt}

\noindent \includegraphics[width=0.5\textwidth]{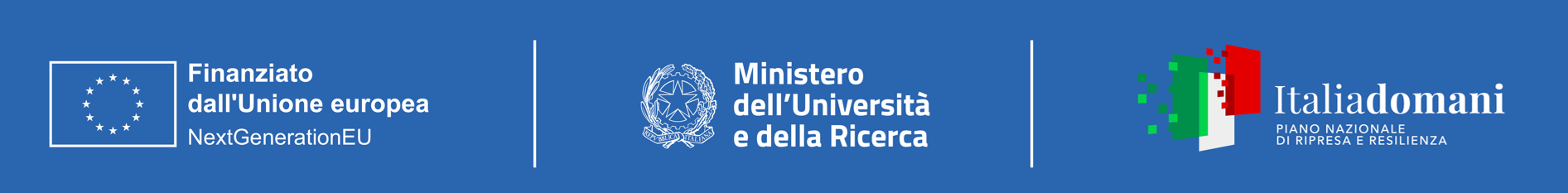}

\bibliographystyle{IEEEtran}

\bibliography{bibtex/bib/refs}

\end{document}